\documentclass[11pt]{article}
\usepackage{paclic32}
\usepackage{times}
\usepackage{latexsym}
\usepackage{amsmath}
\usepackage{multirow}
\usepackage{url}

\usepackage{amsmath,epsfig}
\usepackage{graphicx}
\graphicspath{{50Images/}}
\usepackage{multirow}
\usepackage{url}
\usepackage{soul}
\usepackage{color}

\title{Identifying Computer-Translated Paragraphs using Coherence Features}

\author{Hoang-Quoc Nguyen-Son$^\textsuperscript{1,2}$, Ngoc-Dung T. Tieu$^\textsuperscript{3}$, Huy H. Nguyen$^\textsuperscript{3}$, \\
\textbf{Junichi Yamagishi$^\textsuperscript{1,3,4}$, and Isao Echizen$^\textsuperscript{1,3}$ }\\
  $^\textsuperscript{1}$National Institute of Informatics, Tokyo, Japan \\
    $^\textsuperscript{2}$KDDI Research Inc., Saitama, Japan \\
  $^\textsuperscript{3}$The Graduate University for Advanced Studies, Kanagawa, Japan\\
  $^\textsuperscript{4}$The University of Edinburgh, Edinburgh, United Kingdom\\
  {\tt \{nshquoc,nhhuy,dungtieu,jyamagis,iechizen\}@nii.ac.jp} \\
  }

\date{}

\begin{document}
\maketitle
\begin{abstract}
We have developed a method for extracting the coherence features from a paragraph by matching similar words in its sentences. 
We conducted an experiment with a parallel German corpus containing 2000 human-created and 2000 machine-translated paragraphs. 
The result showed that our method achieved the best performance (accuracy = 72.3\%, equal error rate = 29.8\%) when it is compared with previous methods on various computer-generated text including translation and paper generation (best accuracy = 67.9\%, equal error rate = 32.0\%).
Experiments on Dutch, another rich resource language, and a low resource one (Japanese) attained similar performances.
It demonstrated the efficiency of the coherence features at distinguishing computer-translated from human-created paragraphs on diverse languages.
\end{abstract}

\section{Introduction}
\label{sec:intro}
Computer-translated text plays an essential role in modern life. 
Such artificial text helps people, who use different languages, can communicate each other.
Machine-translated systems thus significantly support or even completely relieve human translators and interpreters from time-consuming burden. 

Thanks to deep learning, neural machine translation have been drastically progressed recently. 
However, we can still tell ``this text must be automatically translated by a computer and reads strangely'' sometimes.
The unexpected quality leads readers to confuse or misunderstand the meaning of artificial text comparing with the original meaning such as machine-translated web pages, especially in low resource languages. 
Enhanced methods are thus needed to identify machine-translated text.

Research on computer-translated text detection has been of interest to the natural language processing community. 
Most detection methods are aimed at the sentence level and use a tree parser~\cite{chae2009predicting,li2015machine} to estimate the naturalness of a text passage. 
However, the scope of these methods is limited to individual sentences and ignores the relationships among sentences. 
In contrast, some methods are aimed at identifying the translation by the POS $N$-gram model~\cite{arase2013machine,aharoni2014automatic}, but they extract features from only a few adjacent words.
Other methods identify the translated documents~\cite{nguyen2017identifying} or generated papers~\cite{labbe2013duplicate} using word distribution.
However, such methods are only suitable for huge text.

Of the various levels of text, i.e., word, phrase, sentence, paragraph and document, the paragraph is one of the most important. 
For instance, paragraphs help readers to quickly receive important content amidst large pieces of text on various topics, such as abstracts of scholarly papers and news summaries. 
Moreover, the paragraph is the main part of current digital text (e.g., email and online product descriptions). 
The paragraph level provides more information compared with the sentence level. 
A large document is often separated into individual paragraphs whose sentences have meanings in common and coherence. 
In this paper, we are interesting in investigating what differentiate of the coherence from machine-translated and human-written paragraphs.
Our hypothesis is that coherence at the paragraph level is one of the factors causing such ``artifacts'' of the computer-generated sentences. 
In order to empirically prove this, we built several classifiers using the coherence features and compare them with ones using previously suggested features.  

\begin{figure*}
\centering
\includegraphics[]{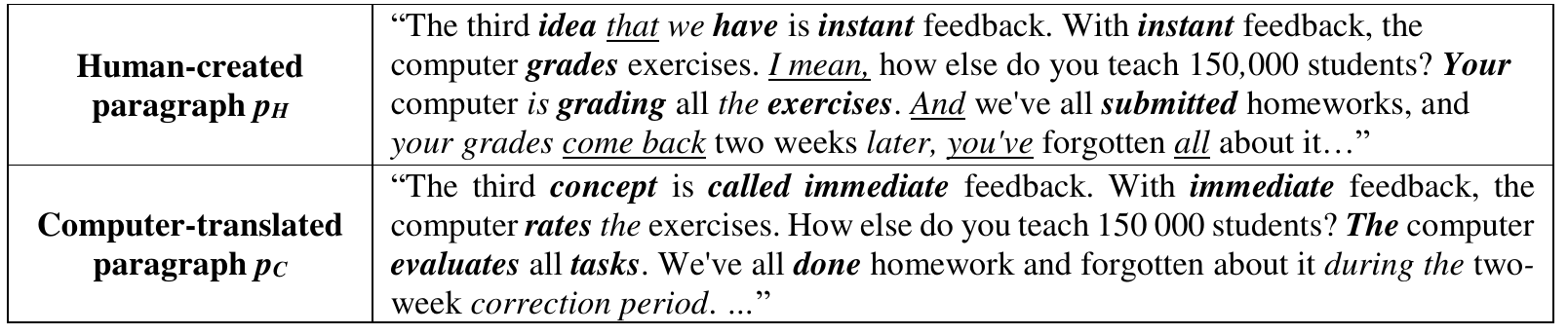}
\caption{Coherence of parallel human-created vs. computer-translated paragraphs. The difference is presented in italic. The using of various similar words is highlighted in bold. The missing words in computer-generated text are described by underline.}
\label{fig:01_CoherenceExamples}
\end{figure*}

A machine-translated text has almost the same meaning and structure as a genuine original one, but the use of words is different. 
Figure~\ref{fig:01_CoherenceExamples} illustrates the difference by italic words.
Translated artificial text tends to be shorter than the original text, as mentioned by \newcite{volansky2013features}. 
A human-created paragraph\footnote{\url{https://www.ted.com/talks/anant_agarwal_why_massively_open_online_courses_still_matter/transcript}} ($p_H$) thus tends to have more supplemental words (the words in underlined) than a parallel computer-translated one ($p_C$).
The translated text also uses different similar words (bold). 
These differences may be the cause of the lower coherence of computer-translated paragraphs. 

In this paper, we present a method for detecting computer-translated text at the paragraph level. 
Our contributions are threefold.

\begin{itemize}

\item 
We propose a metric, $POSMat$, to match similar words for a partial paragraph related to a part of speech (POS) pair.

\item
We present a matching penalty metric, $MatPen$, to reduce the effect of unmatched words. 
The $MatPen$ is integrated with the $POSMat$ into a paragraph coherence metric, $ParaCoh$, to estimate the coherence of a paragraph.

\item
We suggest a method to use the $ParaCoh$ for determining whether a paragraph is translated by a computer.

\end{itemize}

We evaluated the proposed method on 2000 human-transcribed TED paragraphs\footnote{\url{https://www.ted.com}} in English. 
We also collected the corresponding 2000 German human-created paragraphs. 
This text was then translated into English by Google to create parallel machine-generated paragraphs. 
The method-based $ParaCoh$ surpasses previous methods, which identify not only computer-translated but also paper-generated text.
Furthermore, experiments on another rich language (Dutch) produced similar results while a low resource Japanese language achieved even higher performances. 
They demonstrate the capability of the proposed method for recognizing machine-translated text and for evaluating the quality of machine translators.

\section{Related Work}
\label{sec:RelatedWork}

Computer-translated text detection task has been interested by numerous researchers. 
The previous methods are summarized by a taxonomy derived from text granularity falling into sentence, short text, and document.

\subsection{Sentence}
Most primary methods detect computer-generated text at the sentence level on the basis of a parsing tree. For example, \newcite{chae2009predicting} claimed that human-written sentences have a simpler structure than computer-generated ones. The genuine sentence thus contains shorter main phrases including nouns, verbs, adjectives, adverbs, and prepositional phrases. Therefore, they extracted complexity features from the parsing tree, such as parsing depth and average phrase length, in order to distinguish computer-translated from human-written text.

\newcite{li2015machine} also used parsing features to identify artificial translated text. 
They proved that human parsing is more balanced in its structure than machine parsing. 
They extracted balanced-based features from parsing trees such as the ratio of right nodes to left nodes. 
The limitation of parsing-based methods is that they generate parsings only for individual sentences. 
A general tree for multiple sentences (e.g., paragraph, document) cannot be generated. 
Therefore, they cannot quantify relationships among sentences.

\subsection{Short text}

\newcite{arase2013machine} suggested a method to estimate the text fluency of computer-translated text. 
They claimed that translation leads to a ``salad'' phenomenon~\cite{lopez2008statistical}. 
The phenomenon points out the generated text has the same meaning as the original one, but the words in the translation are more chaotic, so it affects the text fluency. 
The authors used a POS $N$-gram language model to quantify the fluency of consecutive words. 
\newcite{nguyen2017detecting} also used a word $N$-gram model to extract features combining with special features, which they called as noise such as misspelled words. 
However, such noise often contains in informal conversations.
Furthermore, \newcite{aharoni2014automatic} extended the POS $N$-gram model by integrate function words features for improving computer-translated identification. 
They argued that automatic translations contain more function words than human-written translations. 
These $N$-gram models only extracted features from a limited number (up to three in common) of adjacent words and ignored the coherence between words separated by one or more other words.

\subsection{Document}

\newcite{nguyen2017identifying} proposed a method to detect a general document using Zipfian law. 
Word distribution is aligned with Zipfian distribution to distinguish computer- with human-generated text. 
The authors proved that human-generated text has more adoption with Zipf's law than machine-generated one. 

On the other hand, \newcite{labbe2013duplicate} detected another kind of computer-produced document, i.e., paper generation.
They pointed out that generated papers contain many duplicated patterns. 
They thus estimated word distribution of a candidate paper with both fake and genuine papers on the basis of inter-textual similarity.
The nearest similarity was used to determine whether a human or a computer creates the input paper.

The restriction of document-based methods is that they need numerous number of words in order to estimate the word distribution. 
Their performances are thus decreased in smaller text.

\section{Proposed method}
\begin{figure*}
\centering
\includegraphics[]{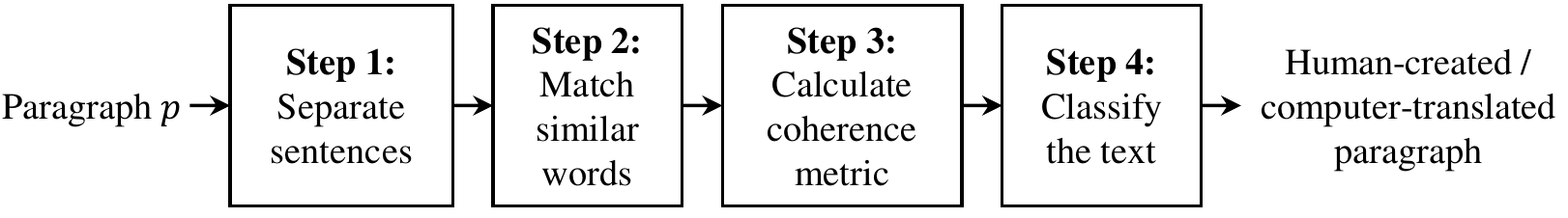}
\caption{Schema for identifying a computer-translated paragraph.}
\label{fig:02_CoherenceScheme}
\end{figure*}

The four steps of the proposed method are illustrated in Figure~\ref{fig:02_CoherenceScheme}.

\begin{itemize}

\item \textbf{Step 1 (\textit{Separate sentences})}: 
The sentences in an input paragraph $p$ are separated using the Stanford CoreNLP splitter~\cite{manning2014coreNLP}.

\item \textbf{Step 2 (\textit{Match similar words})}: 
Each word is matched with another word in the other sentences if they are associated with a POS pair. 

\item \textbf{Step 3 (\textit{Calculate coherence metric})}: 
The similarities of the matched words in each sentence pair are used to calculate the paragraph coherence metric, $ParaCoh$.

\item \textbf{Step 4 (\textit{Classify the text})}: 
$ParaCoh$ is used to determine whether the input $p$ is human-created or computer-translated paragraph.

\end{itemize}

\subsection{Separating Sentences (Step 1)}
The sentences $s_i$ in a paragraph $p$ are separated from the paragraph and placed in set $S$. Separation is done using the Stanford CoreNLP splitter~\cite{manning2014coreNLP}:

\begin{equation}
S = Split(p) = \{s_i\}.
\end{equation}

For example, the first two sentences ($s_{H_1}$ and $s_{H_2}$) in the human-created text example in Figure~\ref{fig:01_CoherenceExamples} are separated from the paragraph:

$s_{H_1}$: ``\textit{The third idea that we have is instant feedback}.''

$s_{H_2}$: ``\textit{With instant feedback, the computer grades exercises}.''

\subsection{Matching Similar Words (Step 2)}
\begin{figure*}
\centering
\includegraphics[]{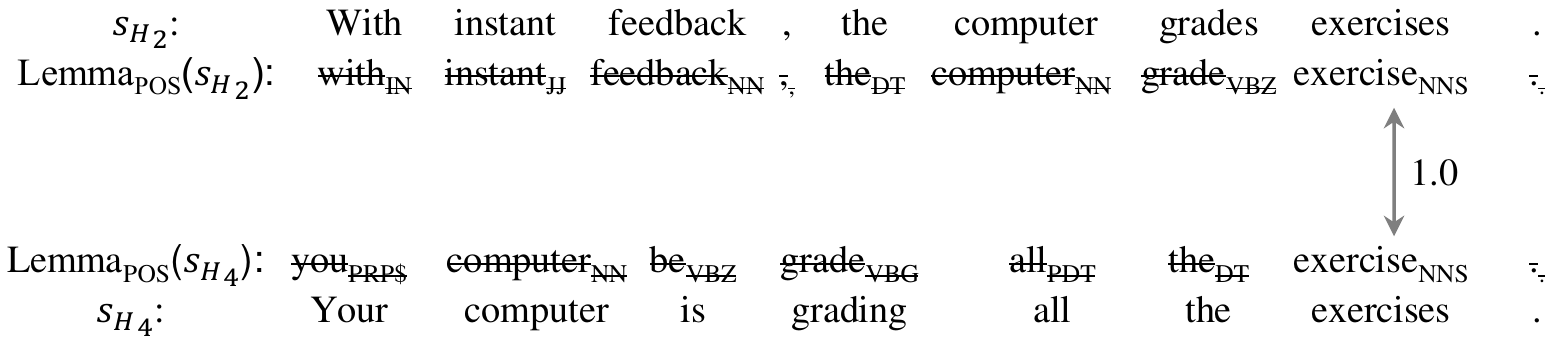}
\caption{Matching plural nouns in human-generated sentence pair.}
\label{fig:03_HumanMatchingExample}
\end{figure*}

English words exist in various grammatical forms which often express similar meaning. For example, a verb ``\textit{be}'' can be represented by different variants (e.g., ``\textit{is}, '' ``\textit{were},'' ``\textit{being},'' ``\textit{'s}''). 
We use the Stanford lemma tool~\cite{manning2014coreNLP} to normalize these words. 
For example, lemmas of several words are shown from the sentence $s_{H_2}$ and $s_{H_4}$ in Figure~\ref{fig:03_HumanMatchingExample}.

A lemma in a sentence is kept if there is another lemma in another sentence and their POSs conform with a processing POS pair while the other lemmas are removed. 
For example, suppose the processing pair consists of two plural nouns \{NNS, NNS\}, and the processing sentences are $s_{H_2}$ and $s_{H_4}$. 
Because both sentences each contain a plural noun, these plural nouns are preserved. 
The other lemmas are eliminated by strikethrough (Figure~\ref{fig:03_HumanMatchingExample}).

A remaining lemma of a sentence is matched with at most one lemma in the other sentence by using the Hungarian algorithm~\cite{kuhn1955hungarian} to ensure that the contributions of the remaining lemmas are balanced. 
The algorithm is also used to match the pairs with maximum similarity in total. 
The similarity of two lemmas is estimated using $path$ metric~\cite{pedersen2004wordnet}. 
This metric is calculated by the shortest path of these lemmas in a Wordnet semantic ontology. 
For example, a plural noun of $s_{H_2}$ is matched with a corresponding plural noun of $s_{H_4}$ and the similarity of these identical lemmas equals 1.0.

\begin{figure*}
\centering
\includegraphics[]{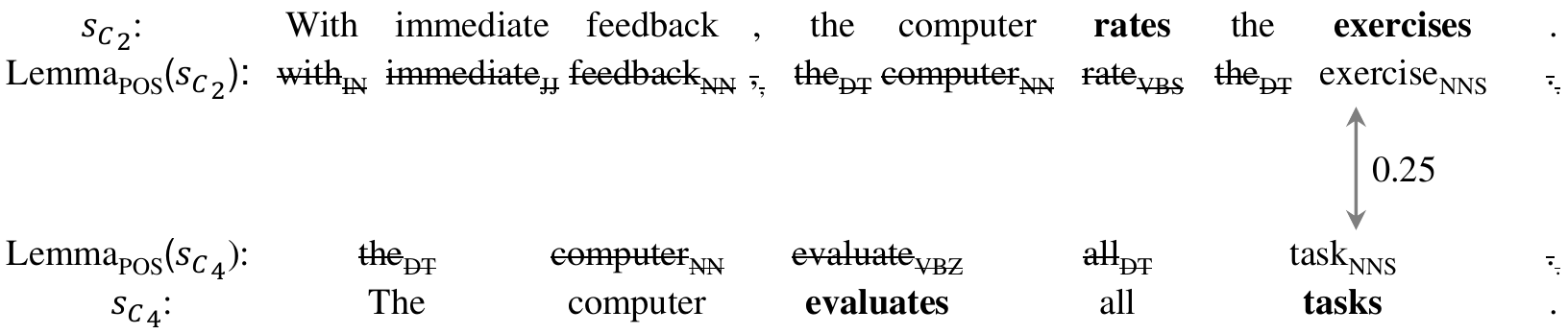}
\caption{Matching plural nouns in machine-generated sentence pair.}
\label{fig:04_MachineMatchingExample}
\end{figure*}

On the other hand, although the meaning of computer-generated text is similar to human-written one (Figure~\ref{fig:01_CoherenceExamples}). 
The use of other similar words effects to text coherence and thus influences to the matching. 
For instance, Figure~\ref{fig:04_MachineMatchingExample} demonstrates the declined matching of two computer-generate sentence $s_{C_2}$ and $s_{C_4}$ due to the use of another word ``\textit{tasks}'' in the second sentence.
Other similar words also result in the matching degradation of other POS pairs, i.e., the use of ``\textit{rates}'' and ``\textit{evaluates}'' (in bold).

\subsection{Calculating Coherence Metric (Step 3)}
The matching words are used to calculate the POS matching metric $POSMat$ for $s_i$ and $s_j$: 

\begin{equation}
POSMat(s_i,s_j)=\dfrac{\sum_{w_k\in s'_i,w_l\in s'_j}path(w_k,w_l )}{n},
\end{equation}

\noindent
where $w_k$ and $w_l$ are pair-matched words for the two sentences, $n$ is the number of matched pairs, $path(w_k,w_l)$ is the $path$ similarity metric of the two matched words estimated using Wordnet~\cite{pedersen2004wordnet} while $s'_i$ and $s'_j$ are two sets which contain remaining words in $s_i$ and $s_j$, respectively. 
For example, the matching metric of $s_{H_2}$ and $s_{H_4}$ is:

\begin{equation}
\begin{split}
POSMat(s_{H_2},s_{H_4})  = \frac{1}{1}=1.
\end{split}
\end{equation}

Since the number of words in $s'_i$ often differs from the number in $s'_j$, we use a penalty matching metric $p$ based on the machine translation METEOR metric~\cite{denkowski2010extending} to reduce the difference:

\begin{equation}
p(s_i,s_j )=0.5\times(\dfrac{|UnMat(s'_i )-UnMat(s'_j )|}{max(|s'_i|,|s'_j|)})^3,
\end{equation}

\noindent
where $UnMat(s')$ is the number of nonmatching words. 
The final matching metric, $POSMat$, is then updated:

\begin{equation}
POSMat(s_i,s_j )=POSMat(s_i,s_j )\times(1-p(s_i,s_j)).
\end{equation}

For example, the matching of $s_{H_2}$ and $s_{H_4}$ is re-estimated using:

\begin{equation}
p(s_{H_2},s_{H_4})=0.5\times(\dfrac{|0-0|}{1})^3=0,
\end{equation}

\begin{equation}
POSMat(s_{H_2},s_{H_4})= 1\times(1-0)=1.
\end{equation}

\begin{figure*}
\centering
\includegraphics[]{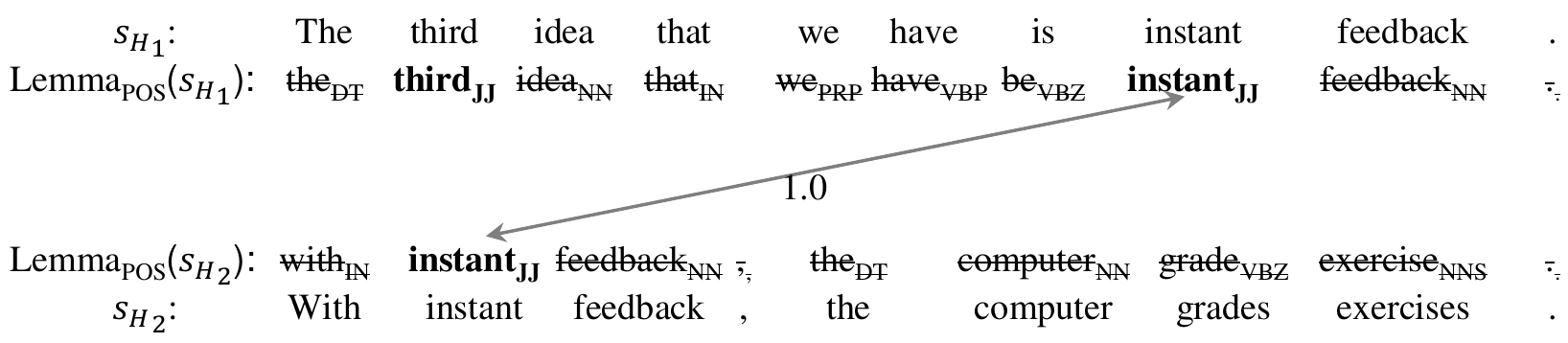}
\caption{Matching similar nouns in human-generated sentence pair.}
\label{fig:05_PenaltyDemontration}
\end{figure*}

Since all candidate words are matched, the $p$ unchanged the similarity matching of $s_{H_2}$ and $s_{H_4}$ related to plural nouns.
Figure~\ref{fig:05_PenaltyDemontration} shows another example of matching adjectives of two human-generated sentences $s_{H_1}$ and $s_{H_2}$.
The $POSMat$ metric of this matching is presented in Equation~\ref{Eq01_penalty_change} demonstrating the effect of nonmatching bold word ``\textit{\textbf{third}}'' into the re-estimated metric.

\begin{equation}
p(s_{H_1},s_{H_2})=0.5\times(\dfrac{|1-0|}{2})^3=0.125,
\end{equation}

\begin{equation}
POSMat(s_{H_1},s_{H_2})= 1\times(1-0.125)=0.875.
\label{Eq01_penalty_change}
\end{equation}

The paragraph coherence metric for a paragraph related to a POS pair is then calculated:

\begin{equation}
\begin{split}
ParaCoh(p)&=\dfrac{\sum_{s_i\in p,s_j\in p,s_i\neq s_j}{POSMat(s_i,s_j)}}{max(1,\binom{m}{2})})\\
&=\dfrac{\sum_{s_i\in p,s_j\in p,s_i\neq s_j}{POSMat(s_i,s_j)}}{max(1,\dfrac{m(m-1)}{2})},
\end{split}
\end{equation}
where $m$ is the number of sentences, the denominator presents for the number possible distinguished sentence pairs in paragraph $p$, and the function $max$ covers a paragraph having only one sentence.

\subsection{Classifying the text (Step 4)}

The coherence features presented by $ParaCoh$ for each POS combination are used to classify human-created and computer-translated text. 
Here, large linear classification (LINEAR)~\cite{fan2008liblinear} outperforms other popular classification algorithms. 
We thus chose LINEAR as the final classifier to determine whether the input paragraph is written by a human or translated by a machine.

\section{Evaluation}
\subsection{Datasets}
We created a dataset from 2100 scripts of recent (2013 to 2018) TED talks\footnote{\url{https://www.ted.com}}. 
These human-created texts were manually transcribed by native English speakers, and then, 2000 paragraphs were randomly extracted. 
The paragraphs contained 14.41 sentences on average.
Then, we collected corresponding 2000 paragraphs translated by native German speakers. 
These paragraphs had the same content as the English ones. 
The German paragraphs were then translated into English by Google Translate to create machine-translated paragraphs. 

\subsection{Comparison with previous methods}
Accuracy (ACC) and equal error rate (EER) was chosen as evaluation metrics, since the corresponding $F$-measures would give equivalent results with the accuracy. 
Four commonly classifiers, which are mentioned in previous methods, including logistic regression (LOGISTIC), support vector machine optimized by stochastic gradient descent (SGD(SVM), SVM optimized by sequential minimal optimization SVM(SMO), LINEAR were run with 10-fold cross validation. 
The performances of previous methods and the proposed method are shown in Table~\ref{tab:01Comparison}.

\begin{table*}
\begin{center}
\begin{tabular}{|c|c c|c c|c c|c c|}

\hline 
\multirow{2}{*}{\textbf{Method}} & \multicolumn{2}{|c|}{\textbf{LOGISTIC}} & \multicolumn{2}{|c|}{\textbf{SGD(SVM)}} & \multicolumn{2}{|c|}{\textbf{SMO(SVM)}} &\multicolumn{2}{|c|}{\textbf{LINEAR}} \\ 
\cline{2-9} 
&\textbf{ACC}&\textbf{EER} &\textbf{ACC} &\textbf{EER} &\textbf{ACC}&\textbf{EER} &\textbf{ACC}&\textbf{EER}\\ 
\hline

\cite{nguyen2017identifying}	&	64.8\%	 &	35.0\%		& \underline{65.0\%}	&	\underline{\textbf{34.8\%}}	& \textbf{65.2\%}	&	35.1\% &	65.0\%	&	34.9\%\\
\cite{li2015machine} &	67.7\%	 &	\textbf{32.5\%}		 & 66.2\%	&	34.2\%	& 67.0\%	&	33.6\%&	\textbf{\underline{67.8\%}}	&	\underline{33.4\%}\\
\cite{aharoni2014automatic}&	67.6\%	 &	32.3\%		 & 66.3\%	&	33.8\%	& \underline{67.4\%}	&	\underline{32.6\%} &	\textbf{67.9\%}	&	\textbf{32.0\%}\\
\hline 
Our&	69.7\%	 &	30.4\%		& 69.8\%	&	30.6\%	&	70.9\%	&	31.1\% &{\color{red} \textbf{72.3\%}}	&	{\color{red}\textbf{29.8\%}}\\
\hline 

\end{tabular}
\caption{Comparison with previous methods on accuracy (ACC) and equal error rate (ERR) metrics. The best metrics are shown in bold. The underline describes for the best classifiers, which are chosen in previous methods. The topmost performance is highlighted by red.}
\label{tab:01Comparison}
\end{center}
\end{table*}

These all methods tended to focus on recognize translation text in different granularity. 
The first method, which was previously used for document-level, compared word distribution with Zipfian one~\cite{nguyen2017identifying}. 
In the second method, \newcite{li2015machine} identified computer-translated text using balancing properties of a parsing tree.
Because a parsing tree is only created for individual sentences, the average of these sentence features was used for the paragraph.
For larger text, another method quantified the frequency of text by using the POS $N$-gram model combining with function words~\cite{aharoni2014automatic}.

As shown in Table~\ref{tab:01Comparison}, the document-based method is degraded their performances on lower granularity because the alignment between word frequency and Zipfician distribution is more effective in huge number of words. 
On the other hand, the sentence-level method using parsing trees were ineffective at the paragraph level as it does not take into account the relationship among sentences in a paragraph.
The extension of $N$-gram features by combining them with function words worked better than the two previous methods, but it only estimates the coherence of consecutive words. 
In contrast, the proposed method overcomes the problem and achieves the best performance across all classifiers.

The experiments of previous methods are given the identical or competitive results with the chosen classifiers in corresponding methods. 
These classifiers are thus used for experiments below. 
With our method, since LINEAR achieved the best performance, it was chosen to create the final classifier.

Table~\ref{tab:02BestPOSPairs} shows the top five performances of 990 in-duplicated POS combination pairs sorted by their accuracies. 
These pairs affect to the coherence machine-translated German text comparing with human-created text.
These pairs should thus be taken more attention in order to improve machine translators in this language.

\begin{table}
\begin{center}
\begin{tabular}{|c|c|c|c|}

\hline 
\textbf{Rank}& \textbf{POS pair} & \textbf{ACC} & \textbf{EER}\\ 
\hline
1 & VB-VBG & 63.7\% & 39.1\%  \\
\hline
2 & VBG-RB & 63.6\% & 39.5\%  \\
\hline
3 & VBG-NN & 63.5\% & 38.5\%  \\
\hline
4 & VBG-VBG & 63.5\% & 40.9\%  \\
\hline
5 & IN-VBG & 62.3\% & 40.3\%  \\
\hline
\end{tabular}
\caption{Performances of top five POS pairs.}
\label{tab:02BestPOSPairs}
\end{center}
\end{table}

\subsection{Other languages}

\begin{figure*}
\centering
\includegraphics[]{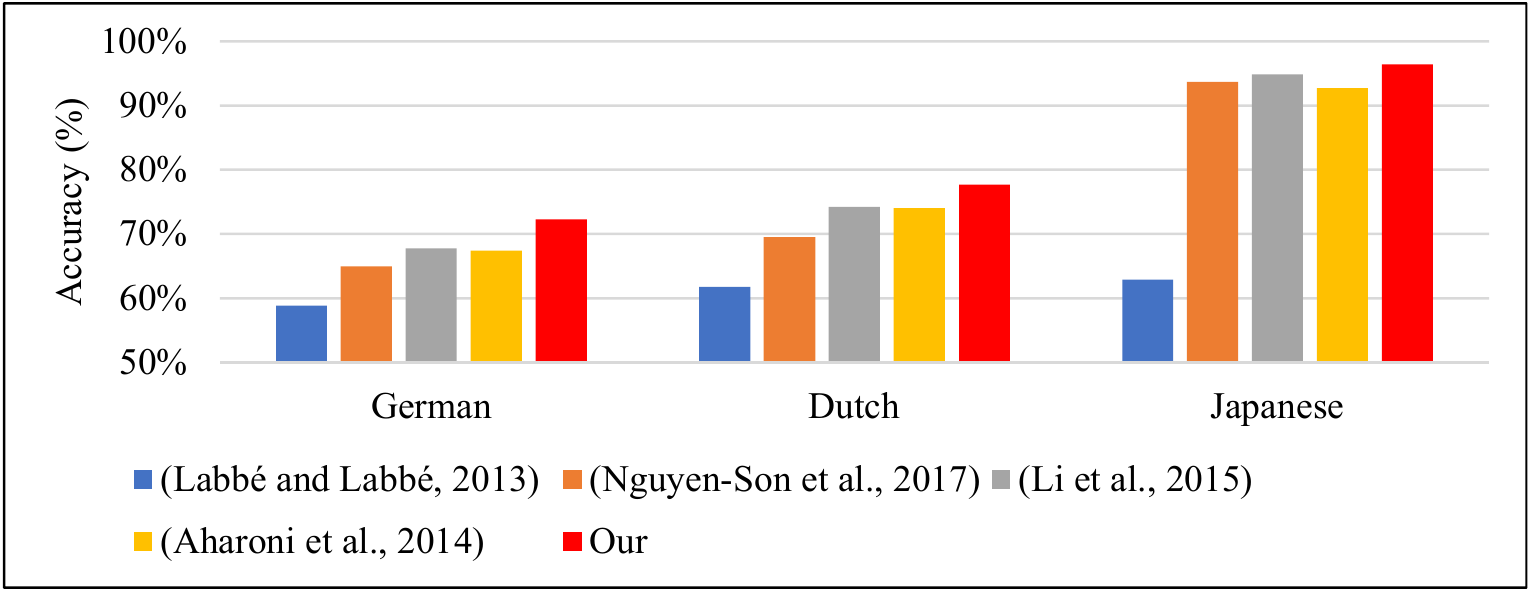}
\caption{Evaluation on various languages.}
\label{fig:06_VariousLanguages}
\end{figure*}

Finally, we conducted similar experiments with another rich language (Dutch) and lower resource one (Japanese).
In each, 2000 human-written and corresponding machine-translated human-written paragraphs are used.
Each paragraph has an average of 14.64 and 14.30 sentences in Dutch and Japanese, respectively.
Since the equal error rate metrics are given similar results above, we only show the accuracy metrics in this comparison. 
Furthermore, we compared with another document-based method~\cite{labbe2013duplicate}, which distinguishes another kind of computer-generated paragraph, i.e., paper generation.
In contrast with other previous methods, which extracted features and uses classifiers for identifying translated-generated paragraph, Labb{\'e} and Labb{\'e} compared word distribution of a candidate document with all training distributions using inter-textual distance. 
The candidate document is labeled as the same type of the nearest comparing.
The result of the comparison is shown in Figure~\ref{fig:06_VariousLanguages}.

The experimental results showed that the paper generation detector achieves the worst performances. 
Because the translation generated the text have almost same meaning with the original text, the word distribution is almost synchronized.
It significantly affects to the hypothesis of the Labb{\'e} and Labb{\'e}'s method through all three languages.
With other methods, two rich resource languages acquired the similar performances. 
On the other hand, experiments on the lower resource language produced significantly better performances. 
Our method reached the best results in the three languages. 
This method can be used to evaluate the quality of translated text in various languages with different resources. 

\section{Conclusion}
The coherence of human-created paragraphs is generally better than that of computer-translated ones. 
The method we propose quantifies the coherence of sentences in a paragraph by matching similar words. 
Our evaluation showed that the coherence features result in higher accuracy than that of state-of-the-art methods on different granularity of German text. 
Moreover, the evaluation of a similar resource language (Dutch) and a low-resource one (Japanese) achieved similar results. 
It also demonstrated another capability of our method on measuring the quality of machine translators on various languages with different resource-levels. 
It significantly supports for current translators recognizing and improving the text generation. 

To the best of our knowledge, current parallel translation corpora support for sentence-level\footnote{\url{http://www.statmt.org/europarl/}}. 
Other larger level datasets use English as original text such as TED talk\footnote{\url{https://www.ted.com/}}.
Our future work targets to create a human translated English dataset from other languages in paragraph level.
Then, we use the coherence features to classify human-translated and computer-translated text.

In another direction, the next research includes using a deep learning network to improve the quantification of the coherence and fluency features.
We also extend the matching algorithm to phrases for further enhancing the coherence metrics. 

\section*{Acknowledgments}
This work was supported by JSPS KAKENHI Grant Numbers JP16H06302 and 18H04120

\bibliographystyle{acl}
\bibliography{ComputerGeneratedTextByMatching}

\end{document}